\documentclass[11pt,a4paper]{style1}

\definecolor{lastauthor}{RGB}{143, 68, 115}

\usepackage{marvosym}

\title{Anti-Persona: Disrupting Unauthorized Identity Binding and Recognition in Personalized Vision--Language Models}

\author[1]{{\large Abhishek Basu$^{*}$}}
\author[1]{{\large Fahad Shamshad$^{*}$}}
\author[1, 2]{{\large Karthik Nandakumar}}
\affiliation[1]{MBZUAI, UAE \hspace{2em}}
\affiliation[2]{Michigan State University, USA}

\usepackage[table]{xcolor} 
 
\newcommand{\myrowcolor}{\rowcolor[HTML]{FFFBE6}}
\usepackage{amsmath}
\usepackage{amssymb}
\usepackage{graphicx}
\usepackage{booktabs}
\usepackage{xcolor}
\usepackage{hyperref}
\usepackage{cleveref}
\usepackage[utf8]{inputenc}   
\usepackage{amsmath}
\usepackage{amssymb}
\usepackage{amsfonts}
\usepackage{booktabs}
\usepackage{multirow}
\usepackage{pifont}
\usepackage{tikz}
\usepackage{pgfplots}
\pgfplotsset{compat=1.17}
\usepackage{algorithm}
\usepackage{algpseudocode}

\crefname{section}{Sec.}{Secs.}
\Crefname{section}{Sec.}{Secs.}
\usepackage{natbib}
\setcitestyle{numbers,square,sort&compress,citesep={,}}
\usepackage{enumitem}
\usepackage{multirow}
\usepackage{microtype}
\usepackage{tabularx}

\usepackage{makecell}

\hypersetup{
  colorlinks=true,
  linkcolor=red,
  citecolor=metablue,
  urlcolor=metablue,
}

\abstract{
Few-shot personalization enables large vision--language models (LVLMs) to learn user-specific visual concepts for applications such as personalized retrieval and subject-aware querying. However, it also creates a privacy risk: an adversary can bind a target identity from a few reference images and subsequently detect that identity in new images through natural-language queries. We introduce Anti-Persona, an image-level defense against unauthorized identity binding and recognition in personalized LVLMs. Our key insight is that identity personalization relies on visual features shared across multiple reference images. We aggregate these features into an identity prototype and optimize visually subtle perturbations that disrupt prototype alignment in the vision-encoder space. Spatial smoothing and low-frequency preservation further promote visual fidelity and practical resilience to image compression. The resulting protection does not depend on a specific prompt and supports both proactive anti-personalization and reactive image protection. Experiments on two representative personalized LVLMs demonstrate protection rates of up to $95.0\%$ while preserving visual fidelity. The method remains stable across prompt variations and evaluated identity-query tasks, and improves black-box transfer under encoder mismatch. 
}

\newcommand{\githubresourceicon}{\includegraphics[width=1.35em]{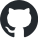}}
\resource{\githubresourceicon}{Code}{https://github.com/iabh1shekbasu/anti-persona}

\begin{document}

\maketitle

\section{Introduction}
\label{sec:intro}
Large Vision-Language Models (LVLMs) have made significant progress in aligning visual and textual representations~\cite{liu2023visual,li2023blip}. Trained on web-scale image-text corpora, they can answer free-form questions, follow visual instructions, and associate language with fine-grained visual concepts that previously required task-specific architectures~\cite{zhang2024vision}. These capabilities have enabled applications such as accessibility tools, creative assistants, and consumer photo platforms~\cite{xu2025personalized}.

An increasingly important capability of LVLMs is \emph{personalization}, which adapts a model to a user-specific visual concept from only a few reference images~\cite{pham2024personalized}. Given several images of a person, a personalized LVLM can learn that individual as a new visual concept and subsequently recognize them in held-out images. This is typically achieved by learning a compact subject-specific representation on top of a frozen backbone~\cite{nguyen2024yo,alaluf2024myvlm}, enabling personalized assistance, photo organization, identity-aware retrieval, and subject-specific querying~\cite{kim2026mmpb}.

While personalization benefits legitimate users, its low cost and few-shot data requirements also make it accessible to adversaries. A few publicly available photos may be sufficient to bind a person's identity to an LVLM~\cite{shi2025pvchat,nguyen2024yo,alaluf2024myvlm}. Once personalized, the model can be queried in natural language to determine whether the target appears in new images~\cite{bai2025online}. This creates a significant privacy risk: \textit{a general-purpose LVLM can be repurposed as a targeted identity recognizer without the person's knowledge or consent}. Unlike conventional face-recognition systems that require dedicated pipelines or specialized models~\cite{zhao2003face,laishram2025toward}, personalized LVLMs expose identity detection through a flexible prompt-based interface, lowering the barrier to unauthorized use  without specialized expertise.

A growing line of work studies image-level protection against unauthorized use, primarily for generative models. Anti-DreamBooth~\cite{van2023anti}, Glaze~\cite{shan2023glaze}, and PhotoGuard~\cite{salman2023raising} apply visually constrained perturbations to interfere with image editing, style imitation, or subject-driven generation. These defenses target synthesis-based threats, where protection is measured by whether a model fails to edit, mimic, or reproduce protected content~\cite{zhao2025visual}. Recognition-based personalization poses a different challenge: the model need not synthesize the target identity, but only learn a discriminative representation sufficient for detecting it. Generative defenses are therefore not directly optimized to prevent identity binding or prompt-based recognition in personalized LVLMs, motivating a protection mechanism tailored to this setting and its distinct privacy requirements. 

\begin{figure*}[t]
    \centering
    \includegraphics[
        width=\textwidth,
        page=1,
        trim={48bp 50bp 53bp 45bp},
        clip
    ]{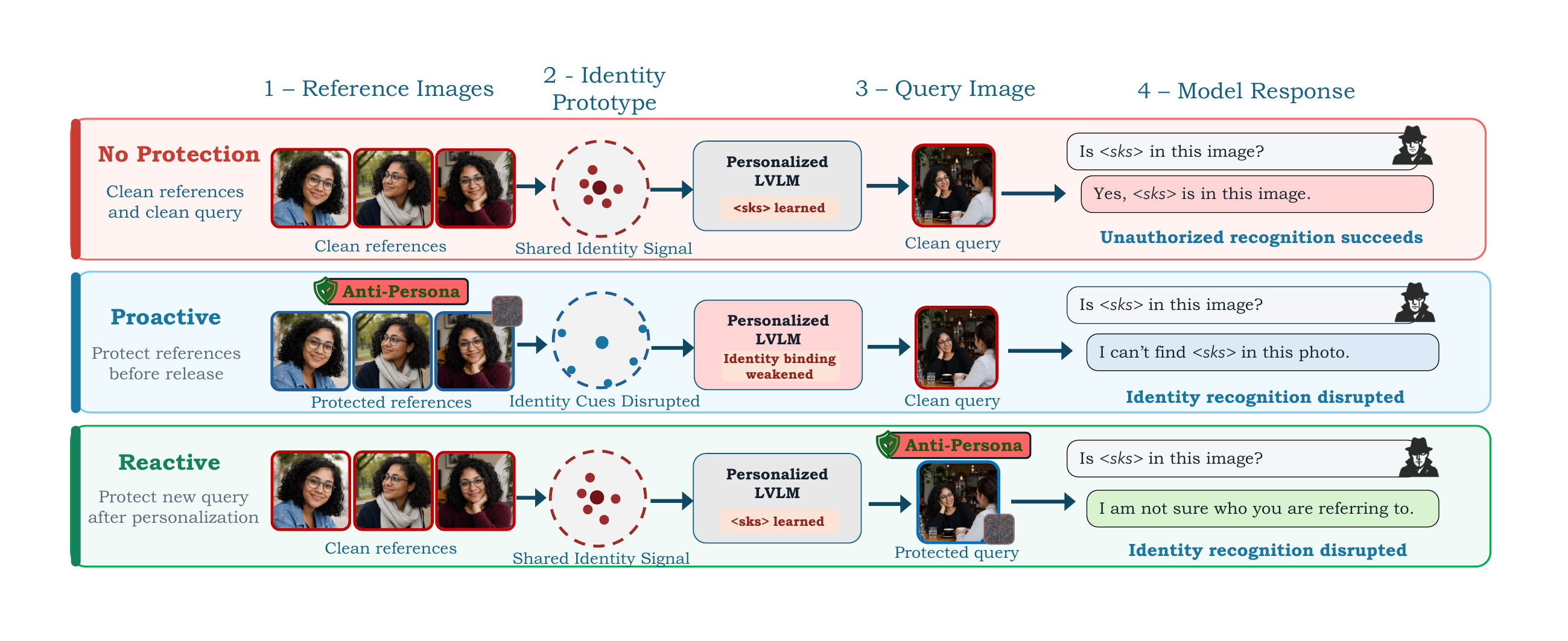}
\caption{\textbf{Anti-Persona protects against unauthorized identity binding
and recognition in LVLMs personalized from only a few reference images.}
Without protection (top), an adversary uses a small set of clean reference
images to bind the target identity to a learned subject representation,
denoted here by $\langle\mathrm{sks}\rangle$. This is an arbitrary identifier
representing the parameters learned specifically for the target individual.
The adversary can then use the personalized model to recognize the individual
in new query images. Anti-Persona constructs an identity prototype by
aggregating consistent identity features across multiple reference images and
optimizes imperceptible perturbations in the vision encoder space to weaken
the alignment between protected image features and this prototype. In the
\emph{proactive} setting (middle), reference images are protected before
release, weakening identity binding and subsequent recognition on clean
queries. In the \emph{reactive} setting (bottom), the LVLM has already been
personalized using clean references. Anti-Persona instead protects newly
released query images, disrupting their match with the learned identity
representation.}
    \label{fig:teaser}
\end{figure*}

In this paper, we introduce an image-level protection framework for recognition-based personalization in LVLMs (see Figure~\ref{fig:teaser}). Our approach follows two key observations. \textit{\textbf{First}}, personalized LVLMs rely on vision-encoder representations as the visual evidence for identity binding. We therefore optimize protective perturbations in the encoder space rather than attacking a particular prompt response. \textit{\textbf{Second}}, identity binding relies on cues that remain stable across multiple images of the same person, rather than on memorizing each image independently. Consequently, perturbing an image only with respect to its own representation may leave shared identity cues available for personalization. We address this limitation by aggregating multiple reference representations into an identity prototype and optimizing each protected image to reduce its alignment with this prototype. This directly targets identity-consistent structure rather than only image-specific features. 

In addition to the identity-disruption objective, we refine the perturbation to make the protection effective under realistic image-sharing conditions. Images shared online are often recompressed, which can attenuate high-frequency adversarial signals. We therefore encourage spatial coherence to suppress abrupt local variations and concentrate perturbation energy in lower-frequency components that are less susceptible to JPEG attenuation~\cite{wallace1991jpeg}. These refinements are designed to improve perceptual quality and compression resilience while preserving the identity-disruption objective. Because the objective operates on identity-relevant representations rather than a specific query, we evaluate it across prompt formulations. We additionally study transfer when the protection encoder differs from that of the target personalized LVLM.

Specifically, our main contributions are:

\begin{itemize}

    \item We identify and formalize a privacy threat in few-shot personalized LVLMs, in which an adversary can bind an individual's identity to an LVLM using only a few reference images and later detect that individual in new images through natural-language queries without the target's consent. We formulate image-level protection for this recognition-based setting, shifting the defense focus from synthesis-oriented misuse in generative models to unauthorized identity binding and recognition.

    \item We propose a multi-reference identity-disruption objective that aggregates encoder representations from multiple reference images into an identity prototype and optimizes protected images to weaken their feature alignment with this prototype. This directly targets identity-consistent cues that recur across multiple reference images rather than image-specific features.

    \item We introduce edge-aware spatial smoothing and low-frequency preservation to reduce visible artifacts and improve resilience to JPEG compression while maintaining the identity-disruption objective.

    \item We evaluate Anti-Persona on representative personalized LVLMs, covering both proactive and reactive protection settings. Our method achieves the strongest overall protection among the evaluated methods while preserving visual fidelity. It remains stable across prompt formulations, extends beyond direct presence queries to VQA-style identity queries, and outperforms the evaluated baselines under strict black-box encoder mismatch.

\end{itemize}

\section{Related Work} \label{sec:related}

\noindent \textbf{Personalization in LVLMs.}
Personalization has been extensively studied in generative models, where methods such as DreamBooth~\cite{ruiz2023dreambooth}, Textual Inversion~\cite{gal2022image}, and Custom Diffusion~\cite{kumari2023multi} adapt text-to-image models to synthesize a specific subject from a few reference images~\cite{zhang2025survey}. In contrast, personalization in LVLMs is more recent and serves a different purpose: rather than generating images of a subject, the model learns to recognize and interact with user-specific visual concepts in downstream vision-language tasks~\cite{pham2024personalized}. Recent methods such as Yo'LLaVA~\cite{nguyen2024yo} adopt token-based personalization, where a learned identifier is bound to a target subject, while MyVLM~\cite{alaluf2024myvlm} introduces architectural components such as a dedicated concept head for personalized recognition. These approaches show that LVLMs can acquire identity-specific representations from only a small number of examples, enabling applications such as personalized retrieval and subject-aware querying. \textit{However, this capability also exposes new vulnerabilities, as the same mechanisms that enable efficient personalization can be exploited to inject sensitive identities into models without user consent}.

\noindent \textbf{Protection against Unauthorized Personalization.}
A growing body of work has explored adversarial perturbations as a defense against unauthorized use of personal images by generative AI systems~\cite{ma2026safety}. General image protection methods such as PhotoGuard~\cite{salman2023raising} and Glaze~\cite{shan2023glaze} perturb images to resist malicious image editing or style imitation~\cite{ye2025evaluating}, while methods such as Anti-DreamBooth~\cite{van2023anti}, MIST~\cite{liang2023mist}, Anti-Diffusion~\cite{zheng2025anti}, and AdvDM~\cite{liang2023adversarial} specifically target subject-driven personalization in diffusion models. These approaches are designed for synthesis-based threats, where protection is achieved by preventing a generative model from faithfully editing, mimicking, or reproducing protected visual content. Recognition-based personalization in LVLMs presents a different defense objective. The personalized model need not synthesize the target identity; it only needs to learn a discriminative representation that supports prompt-based identity detection. As a result, defenses designed for generative misuse are not directly optimized to prevent identity binding or subsequent recognition in personalized LVLMs. \textit{In this work, we address this gap by formulating image-level protection for recognition-based personalization, where the goal is to disrupt the identity-consistent features used by personalized LVLMs for detection}.

\section{Preliminaries}

\subsection{Personalization in LVLMs}

A large vision--language model (LVLM) can be represented as a vision encoder followed by a language model. Given an image $x \in \mathbb{R}^{H \times W \times 3}$, a frozen vision encoder $f_\phi$ produces a visual representation $f_\phi(x) \in \mathbb{R}^d$, where any visual projection module is absorbed into $f_\phi$. Given a user query $q$, the language model produces a response as $g_\theta(f_\phi(x),q)$. Although this model can answer general visual questions, it does not explicitly bind new user-specific identities absent from its training data.

Personalization adapts the LVLM to a subject $s$ using a small reference set $\mathcal{R}_s=\{x_1,x_2,\ldots,x_n\}$, whose images depict the same individual under different visual conditions. The systems considered in this work keep the pretrained vision--language backbone frozen and learn lightweight subject-specific parameters $\psi_s$ from $\mathcal{R}_s$. After personalization, the response to a test image $x_{\text{test}}$ and query $q$ depends on the test-image representation $f_\phi(x_{\text{test}})$ and $\psi_s$. These parameters may be implemented as identifier tokens, lightweight prediction heads, or related personalization modules.

Representative personalized LVLMs instantiate $\psi_s$ differently. Yo'LLaVA~\cite{nguyen2024yo} learns latent identifier tokens, including a subject token such as $\langle\texttt{sks}\rangle$, using conversational supervision. MyVLM~\cite{alaluf2024myvlm} instead trains a lightweight concept head $h_{\psi_s}:\mathbb{R}^d\rightarrow\mathbb{R}$ using positive target images and negative or distractor images. Despite these architectural differences, both methods learn subject-specific parameters from encoder representations, making the vision encoder a common interface for identity personalization. Consequently, if protected images no longer produce a consistent identity signal in the encoder space, the personalization mechanism receives weaker evidence for binding or recognizing the target subject. We therefore formulate protection at the encoder-representation level rather than against a specific prompt, response, or personalization head, allowing the defense to target identity evidence shared across different personalized LVLM designs.

\subsection{Threat Model}
\label{sec:threat}

We study unauthorized identity recognition using personalized LVLMs. Let $s$ denote the target individual. The adversary seeks to determine whether $s$ appears in a query image, whereas the defender releases protected images $\tilde{x}=x+\delta$ that disrupt identity binding or recognition while remaining visually similar to $x$.

\noindent\textbf{Adversary goals and capabilities.}
Given reference images of $s$, the adversary follows the standard personalization pipeline of the target LVLM: the pretrained vision--language backbone remains frozen, and subject-specific parameters $\psi_s$ are learned from the available references. The adversary then uses natural-language prompts to determine whether $s$ appears in new images and may vary the formulation of these prompts. We consider a non-adaptive adversary using an off-the-shelf personalized LVLM; modifying the personalization algorithm, purification, adversarial training, and explicit perturbation removal are outside our scope.

\noindent\textbf{Defender goals and capabilities.}
The defender aims to prevent reliable recognition of $s$ while preserving visual fidelity. The defender uses the user's own reference images to estimate the identity prototype but does not assume knowledge of the adversary's prompts. In the \emph{gray-box} setting, the defender has access to the target vision encoder $f_\phi$. In the \emph{black-box} setting, the defender has no access to the target encoder's parameters or gradients and instead optimizes using a surrogate encoder $f_{\phi'}$.

\noindent\textbf{Protection protocols.}
We consider two protocols that differ in when protection is applied. In the \emph{proactive} protocol, the defender protects reference images before release, producing $\tilde{\mathcal{R}}_s=\{\tilde{x}_1,\ldots,\tilde{x}_n\}$. The adversary subsequently personalizes the LVLM using $\tilde{\mathcal{R}}_s$ to obtain $\tilde{\psi}_s$ and evaluates it on clean held-out images. In the \emph{reactive} protocol, the adversary has already learned $\psi_s^{\mathrm{clean}}$ from clean references $\mathcal{R}_s^{\mathrm{clean}}$; the defender instead protects newly released query images before they are fed to the personalized model.

\noindent\textbf{Protection requirements.}
Effective protection must prevent reliable recognition of $s$ under the corresponding protocol: identity binding and subsequent recognition should fail after personalization with protected references in the proactive protocol, whereas protected queries should not be recognized by a model personalized with clean references in the reactive protocol. Protection should remain effective across the prompt variations and encoder-access regimes while preserving visual similarity to the original images.

\begin{figure*}[t]
    \centering
    \includegraphics[width=1\textwidth]{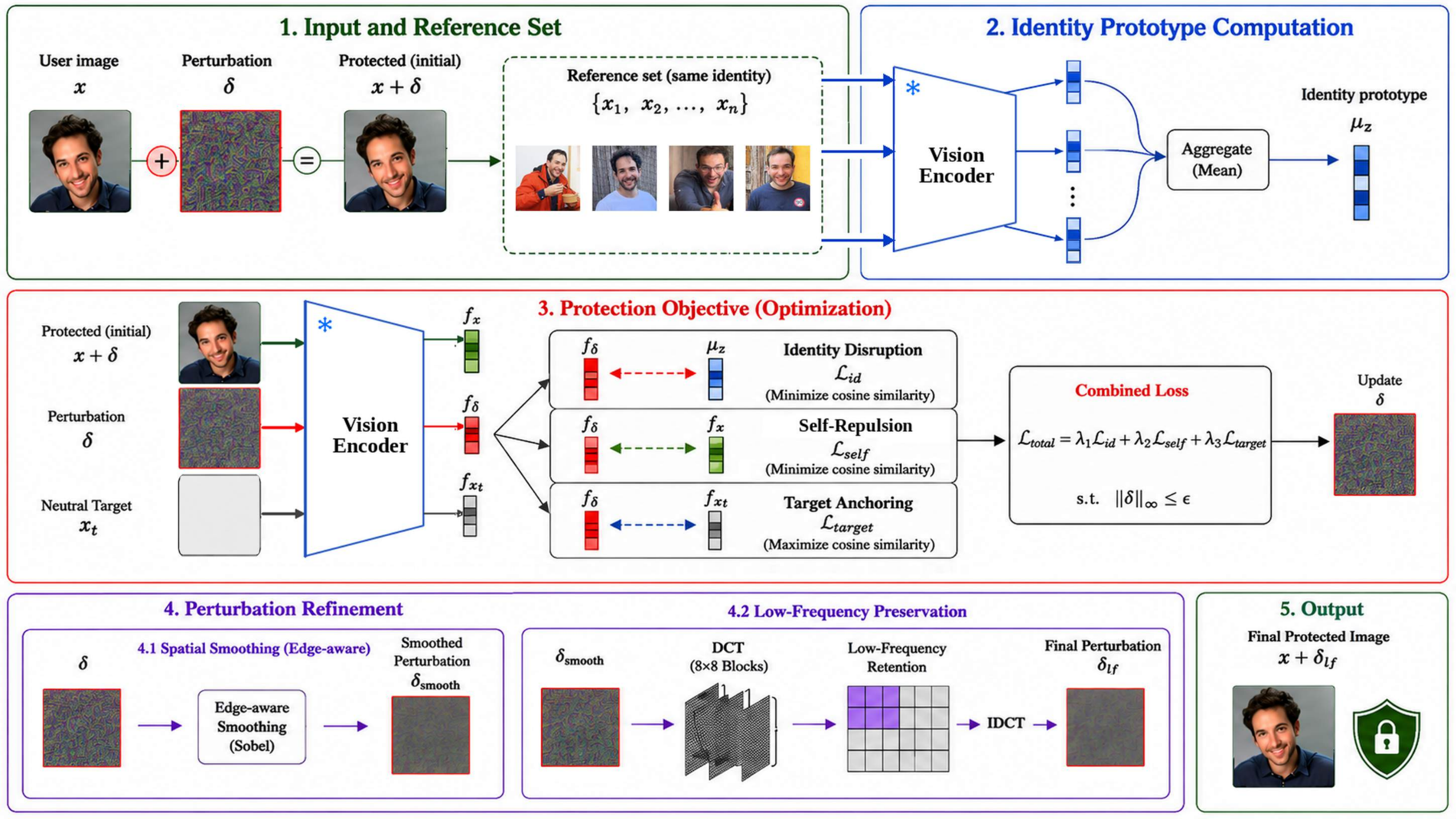}
\caption{
Overview of the proposed identity-prototype protection framework. 
Given a user image and a reference set of the same identity, we first compute an identity prototype by aggregating encoder representations across reference images. 
The protected image is then optimized in the vision-encoder space using a combined loss that disrupts alignment with the identity prototype, reduces residual similarity to the clean image, and anchors the representation toward a neutral target. 
Finally, spatial smoothing and low-frequency preservation refine the perturbation to improve visual quality and robustness to compression, producing a final protected image that is difficult to bind or recognize by personalized LVLMs.
}
\label{fig:method}
\end{figure*}

\section{Method}
\label{sec:method}

Our goal is to construct an imperceptible perturbation $\delta$ for a user image $x$ such that the protected image $\tilde{x}=x+\delta$ is difficult to associate with the target identity $s$ by a personalized LVLM. We formulate protection in the representation space of the vision encoder, since this is the shared interface through which visual identity information is provided to downstream personalization modules. The design follows three principles. \textit{\textbf{First}}, the perturbation should disrupt identity-relevant encoder features before they are used for personalization or recognition. \textit{\textbf{Second}}, because identity personalization is learned from multiple reference images, protection should target the identity-consistent structure shared across those images rather than only image-specific features. \textit{\textbf{Third}}, the perturbation should remain visually unobtrusive and robust to realistic image processing, particularly lossy compression. We first define an identity prototype that captures the shared identity signal, and then use it to build the protection objective. The overall framework is illustrated in Figure~\ref{fig:method}. 

\subsection{Identity Prototype}

A naive way to perturb an image is to move its vision encoder representation away from its clean embedding, for example by reducing the similarity between $f_\phi(x+\delta)$ and $f_\phi(x)$. This instance-level objective can suppress features of the individual image, but it does not explicitly target the information that supports few-shot identity personalization. In this setting, the personalized model observes multiple reference images of the same individual and can rely on cues that remain stable across variations in pose, illumination, expression, and background. As a result, perturbing each image only with respect to its own embedding may leave the shared identity signal sufficiently intact for the model to bind or recognize the subject.

To target this shared signal, we define an identity prototype $\mu_s$ from the user's reference set $\mathcal{R}_s=\{x_1,x_2,\ldots,x_n\}$. Let $z_i=f_\phi(x_i)$ denote the encoder representation of the $i$-th reference image. We compute the prototype as the centroid of the reference embeddings:
\begin{equation}
\mu_s = \frac{1}{n}\sum_{i=1}^{n} z_i
= \frac{1}{n}\sum_{i=1}^{n} f_\phi(x_i).
\label{eq:prototype}
\end{equation}

The prototype captures identity-consistent features shared across the reference images while averaging out variations specific to individual samples, such as pose, lighting, or background. As a result, $\mu_s$ provides a stable representation of the subject's identity in the encoder feature space and serves as the natural target for perturbation. Optimizing protected images to move away from $\mu_s$, rather than from their individual embeddings, ensures that the perturbations disrupt features that are consistent across the identity, yielding stronger and more reliable protection.

\begin{algorithm}[t]
\caption{Identity-Prototype Protection}
\label{alg:protection}
\begin{algorithmic}[1]
\Require Image to protect $x$; reference set $\mathcal{R}_s=\{x_1,\dots,x_n\}$;
frozen encoder $f_\phi$; neutral anchor $x_t$; perturbation budget $\epsilon$;
step size $\alpha$; steps $T$; weights $\lambda_1,\lambda_2$;
Gaussian smoother $G(\cdot)$; low-pass mask $M_{\mathrm{lp}}$
\Ensure Protected image $\tilde{x}$
\Statex
\State $\mu_s \gets \dfrac{1}{n}\sum_{i=1}^{n} f_\phi(x_i)$
\Comment{identity prototype, Eq.~\ref{eq:prototype}}
\State $M_{\mathrm{edge}} \gets \textsc{Sobel}(x)$
\Comment{edge mask from the clean image}
\State initialize $\delta \gets 0$
\For{$t = 1$ \textbf{to} $T$}
    \State $\tilde{x} \gets \mathrm{clip}_{[0,1]}(x+\delta)$
    \State $\mathcal{L} \gets
        \cos\!\big(f_\phi(\tilde{x}),\mu_s\big)
        + \lambda_1 \cos\!\big(f_\phi(\tilde{x}),f_\phi(x)\big)
        + \lambda_2\!\left(1-\cos\!\big(f_\phi(\tilde{x}),f_\phi(x_t)\big)\right)$
        \Comment{Eq.~\ref{eq:objective}}
    \State $\delta \gets \delta - \alpha\,\mathrm{sign}\!\left(\nabla_{\delta}\mathcal{L}\right)$
    \Comment{PGD descent step}
    \State $\delta \gets \mathrm{clip}_{[-\epsilon,\epsilon]}(\delta)$
    \Comment{$\ell_\infty$ projection}
    
    \Statex \hspace{\algorithmicindent}\textit{// perturbation refinement}
    
    \State $\delta \gets G(\delta)\odot M_{\mathrm{edge}} + \delta\odot(1-M_{\mathrm{edge}})$
    \Comment{spatial smoothing}
    \State $\delta \gets \mathrm{IDCT}\!\left(\mathrm{DCT}(\delta)\odot M_{\mathrm{lp}}\right)$
    \Comment{low-frequency preservation}
    \State $\delta \gets \mathrm{clip}_{[-\epsilon,\epsilon]}(\delta)$
    \Comment{re-project}
\EndFor
\State \Return $\mathrm{clip}_{[0,1]}(x+\delta)$
\end{algorithmic}
\end{algorithm}

\subsection{Protection Objective}

Given an image $x$ of subject $s$, we optimize a perturbation $\delta$ such that the protected image $\tilde{x}=x+\delta$ has weak alignment with the identity evidence used by personalized LVLMs. The objective combines three complementary terms: a \textit{prototype-level identity} term that disrupts the shared identity prototype, an \textit{instance-level} term that suppresses residual similarity to the clean image, and a \textit{target-anchoring} term that stabilizes the optimization in feature space. The proposed approach is summarized in Algorithm~\ref{alg:protection}.

\noindent \textbf{Prototype-level identity disruption.}
We first reduce the alignment between the protected embedding and the identity prototype $\mu_s$ defined in Eq.~\ref{eq:prototype}. Since CLIP-based encoders commonly use normalized features and cosine similarity, we minimize
\begin{equation}
\mathcal{L}_{\mathrm{id}}(\tilde{x}) =
\cos\big(f_\phi(\tilde{x}), \mu_s\big),
\end{equation}
which directly penalizes alignment with the identity-consistent direction in the encoder space.

\noindent \textbf{Instance-level repulsion.}
Although the prototype term targets shared identity structure, the protected image may still remain close to its original embedding and retain image-specific identity cues. We therefore introduce an instance-level repulsion term,
\begin{equation}
\mathcal{L}_{\mathrm{self}}(\tilde{x}) =
\cos\big(f_\phi(\tilde{x}), f_\phi(x)\big),
\end{equation}
which reduces similarity between the protected and clean representations. This term complements the prototype loss by suppressing residual cues tied to the individual image.

\noindent \textbf{Target anchoring.}
Untargeted perturbations can lead to unstable optimization when the representation moves away from the identity region. To stabilize this process, we introduce a fixed target image $x_t$ and encourage alignment with its embedding:
\begin{equation}
\mathcal{L}_{\text{target}}(\tilde{x}) = 1 - \cos\big(f_\phi(\tilde{x}), f_\phi(x_t)\big),
\end{equation}
which provides a consistent optimization direction. This formulation is consistent with prior observations that targeted adversarial objectives provide a more stable and consistent optimization direction compared to purely untargeted objectives~\cite{zheng2023targeted}. We use a gray image as $x_t$, serving as a neutral anchor that does not correspond to any specific identity.

The final protection objective is
\begin{equation}
\min_{\delta} \ 
\mathcal{L}_{\mathrm{id}}(\tilde{x})
+ \lambda_1 \mathcal{L}_{\mathrm{self}}(\tilde{x})
+ \lambda_2 \mathcal{L}_{\mathrm{target}}(\tilde{x})
\quad
\text{s.t.}
\quad
\|\delta\|_\infty \leq \epsilon,
\label{eq:objective}
\end{equation}
where $\lambda_1$ and $\lambda_2$ balance the loss terms. We minimize this objective by projected gradient descent under the $\ell_\infty$ constraint, which enforces imperceptibility.

\subsection{Perturbation Refinement}

The objective in Eq.~\ref{eq:objective} disrupts identity representations, but it does not control the structure of the perturbation in image space. This can limit practical deployment in two ways. \textit{First}, an $\ell_\infty$ constraint bounds the perturbation magnitude but does not prevent locally irregular patterns that may become perceptible. \textit{Second}, adversarial signals concentrated in high-frequency components can be weakened by lossy image compression, which is common in online sharing pipelines. We address these issues with two refinements applied after each optimization step.

\noindent \textbf{Spatial smoothing.}
To reduce perceptible artifacts, we regularize the perturbation by suppressing abrupt local variations. Specifically, we apply Gaussian smoothing $G(\cdot)$ using a Sobel-derived mask $M_{\mathrm{edge}}$ computed from the clean image:
\[
\delta \leftarrow G(\delta) \odot M_{\mathrm{edge}} + \delta \odot (1-M_{\mathrm{edge}}).
\]
This operation encourages spatial coherence in visually sensitive regions while preserving the overall direction of the optimized perturbation. As a result, the perturbation remains aligned with the identity-disruption objective while reducing local artifacts that may otherwise be visible.

\noindent \textbf{Low-frequency preservation.}
To improve robustness to compression, we further constrain the perturbation to retain low-frequency structure. After each update, we apply a patch-wise discrete cosine transform (DCT) to $\delta$, keep coefficients selected by a low-pass mask, and reconstruct the perturbation using the inverse DCT. This concentrates perturbation energy in spectral components that are more likely to survive JPEG compression, rather than in high-frequency components that are easily attenuated. The refinement therefore improves compression robustness while preserving the identity-disruption objective.

\section{Experiments} \label{sec:experiments} 
\subsection{Experimental Setup} 
We evaluate whether the proposed protection reduces unauthorized identity recognition while preserving visual fidelity. Our experiments cover reactive and proactive protection, two personalized LVLM architectures, prompt and query variation, black-box encoder transfer, and sensitivity to the number of reference images. 

\noindent\textbf{Implementation details.} We implement the method in PyTorch and generate protected images at $336\times336$ resolution. Unless otherwise stated, pixel-space perturbations use an $\ell_\infty$ budget of $\epsilon=4/255$ and are clipped to the valid image range after every update. Spatial smoothing uses a Gaussian kernel with $(\sigma,k)=(0.8,5)$. In the \emph{gray-box} setting, protection is optimized using the victim-aligned OpenAI CLIP ViT-L/14@336 encoder~\cite{radford2021learning}. In the \emph{black-box} setting, this encoder is never used. Instead, we form a six-encoder surrogate pool comprising CLIP-B/16, CLIP-L/14@224, CLIPA-L/14@336, SigLIP~\cite{zhai2023sigmoid}, ConvNeXtV2~\cite{woo2023convnext}, and supervised ViT-L. Each leave-one-surrogate-out configuration optimizes over the remaining five encoders and is evaluated on Yo'LLaVA. Experiments use a single NVIDIA RTX A6000 (48GB) GPU. 

\noindent\textbf{Dataset and personalized LVLMs.} We use the personalized identity-recognition benchmark from Yo'LLaVA~\cite{nguyen2024yo}, which contains ten identities with approximately 15--20 images per identity. Following the benchmark protocol, 5--10 images are used as personalization references and the remaining valid images are held out for evaluation. We evaluate Yo'LLaVA~\cite{nguyen2024yo}, which learns subject-specific identifier tokens, and MyVLM~\cite{alaluf2024myvlm}, which trains a lightweight concept head over frozen visual features. These systems represent two distinct encoder-based personalization mechanisms. 

\noindent\textbf{Baselines.} We report an unprotected (\emph{Clean}) reference and compare against two baseline families. First, we consider five $\ell_\infty$-bounded pixel-space attacks: FGSM~\cite{goodfellow2015explaining}, I-FGSM~\cite{kurakin2017adversarial}, MI-FGSM~\cite{dong2018boosting}, TI-FGSM~\cite{dong2019evading}, and PGD~\cite{madry2018towards}. Each baseline optimizes an instance-level feature-repulsion objective that reduces the similarity between $f_\phi(\tilde{x})$ and $f_\phi(x)$, providing a controlled comparison with conventional feature-space perturbations. These baselines use the same perturbation budget and encoder-access regime as ours. For broader comparison, we also report DiffAttack~\cite{chen2024diffattack} and WDP~\cite{salar2025enhancing}, two unrestricted diffusion-based references. Because they are not $\ell_\infty$-bounded, we distinguish them from equal-budget comparisons and report their visual fidelity. Within each protocol, all evaluated methods use the same identities, personalized Large VLMs, image resolution, and text prompts. 

\noindent\textbf{Evaluation protocols.} We evaluate protection under two
protocols corresponding to the threat model. Following Sec.~\ref{sec:threat}, the \emph{reactive} protocol personalizes the LVLM using clean references and evaluates it on protected held-out images. The \emph{proactive} protocol instead personalizes the model using protected references and evaluates it on clean held-out images. More details are provided in suppl. material.

\noindent\textbf{Evaluation metrics.} For each target-present query, let $r_i=1$ if a fixed, method-independent parser determines that the response recognizes the target subject, and $r_i=0$ otherwise. We define \[ \operatorname{ProtectionRate} = \frac{100}{N}\sum_{i=1}^{N}(1-r_i), \] where $N$ is the number of evaluated queries. Higher values indicate more frequent failures to recognize the target. The same parser is used for every method and prompt. We report PSNR and SSIM~\cite{wang2004image} between clean and protected images to quantify visual fidelity.

\subsection{Results}

\subsubsection{Main Protection Results}

\begin{table}[t]
    \centering
    \caption{Reactive protection. Protection rate (PR) is the identity-recognition failure rate. Pixel-space methods use $\epsilon=4/255$; higher is better for all metrics.}
    \label{tab:image_protection}
    \setlength{\tabcolsep}{3pt}
    \renewcommand{\arraystretch}{1}
    \begin{tabular}{@{}lcccc@{}}
        \toprule
        \multirow{2}{*}{Method}
        & \multicolumn{2}{c}{PR (\%) $\uparrow$}
        & \multirow{2}{*}{PSNR $\uparrow$}
        & \multirow{2}{*}{SSIM $\uparrow$} \\
        \cmidrule(lr){2-3}
        & Yo'LLaVA & MyVLM & & \\
        \midrule
        Clean       & 3.3  & 50.0 & --    & --    \\
        \midrule
        FGSM        & 2.2  & 60.0 & 36.05 & 0.900 \\
        I-FGSM      & 87.8 & 58.9 & 37.94 & 0.940 \\
        MI-FGSM     & 80.0 & 54.4 & 36.19 & 0.900 \\
        TI-FGSM     & 40.0 & 54.4 & 37.72 & 0.950 \\
        PGD         & 85.6 & 58.9 & 37.92 & 0.940 \\
        DiffAttack & 5.6  & 62.2 & 27.85 & 0.831 \\
        WDP       & 14.4 & 65.6 & 26.77 & 0.817 \\
        \midrule
        \myrowcolor Ours
        & 93.3 & \textbf{71.1} & \textbf{39.30} & \textbf{0.958} \\
        \bottomrule
    \end{tabular}
\end{table}

\begin{table*}[t]
    \centering
    \caption{
        Proactive protection on Yo'LLaVA. The model is personalized
        with protected references and evaluated on clean queries.
        Values are protection rates (\%); higher is better.
        Pixel methods use $\epsilon=4/255$. This protocol evaluates
        whether protection disrupts identity binding during
        personalization rather than only causing evasion at inference.
    }
    \label{tab:proactive_results}

    \setlength{\tabcolsep}{2.8pt}
    \renewcommand{\arraystretch}{1.05}
    \resizebox{\textwidth}{!}{%
        \begin{tabular}{@{}l*{11}{c}@{}}
            \toprule
            \textbf{Method}
            & \textbf{Ciin}
            & \shortstack{\textbf{Denis}\\\textbf{dang}}
            & \textbf{Khanhvy}
            & \textbf{Oong}
            & \shortstack{\textbf{Phuc}\\\textbf{map}}
            & \textbf{Thao}
            & \textbf{Thuytien}
            & \textbf{Viruss}
            & \shortstack{\textbf{Willin}\\\textbf{vietnam}}
            & \textbf{Yuheng}
            & \textbf{Avg.} \\
            \midrule
            Clean
            & 0.0 & 0.0 & 0.0 & 0.0 & 0.0 & 0.0
            & 0.0 & 0.0 & 10.0 & 40.0 & 5.0 \\
            \midrule
            FGSM
            & 60.0 & 30.0 & 100.0 & 10.0 & 0.0 & 100.0
            & 100.0 & 60.0 & 10.0 & 80.0 & 55.0 \\
            MI-FGSM
            & 100.0 & 70.0 & 80.0 & 80.0 & 30.0 & 100.0
            & 100.0 & 90.0 & 0.0 & 80.0 & 73.0 \\
            TI-FGSM
            & 100.0 & 90.0 & 100.0 & 100.0 & 60.0 & 100.0
            & 100.0 & 40.0 & 80.0 & 100.0 & 87.0 \\
            PGD
            & 20.0 & 90.0 & 57.1 & 20.0 & 50.0 & 100.0
            & 100.0 & 100.0 & 40.0 & 100.0 & 67.7 \\
            DiffAttack
            & 60.0 & 60.0 & 100.0 & 40.0 & 80.0 & 100.0
            & 80.0 & 100.0 & 0.0 & 80.0 & 70.0 \\
            WDP
            & 100.0 & 20.0 & 100.0 & 100.0 & 20.0 & 100.0
            & 40.0 & 0.0 & 70.0 & 100.0 & 65.0 \\
            \midrule
            \myrowcolor \textbf{Ours}
            & 100.0 & 100.0 & 100.0 & 100.0 & 70.0 & 100.0
            & 80.0 & 100.0 & 100.0 & 100.0 & \textbf{95.0} \\
            \bottomrule
        \end{tabular}%
    }
\end{table*}

\begin{figure}[t]
    \centering
    \includegraphics[
        width=0.5\linewidth,
        trim=6.5cm 7cm 5.8cm 0.5cm,
        clip
    ]{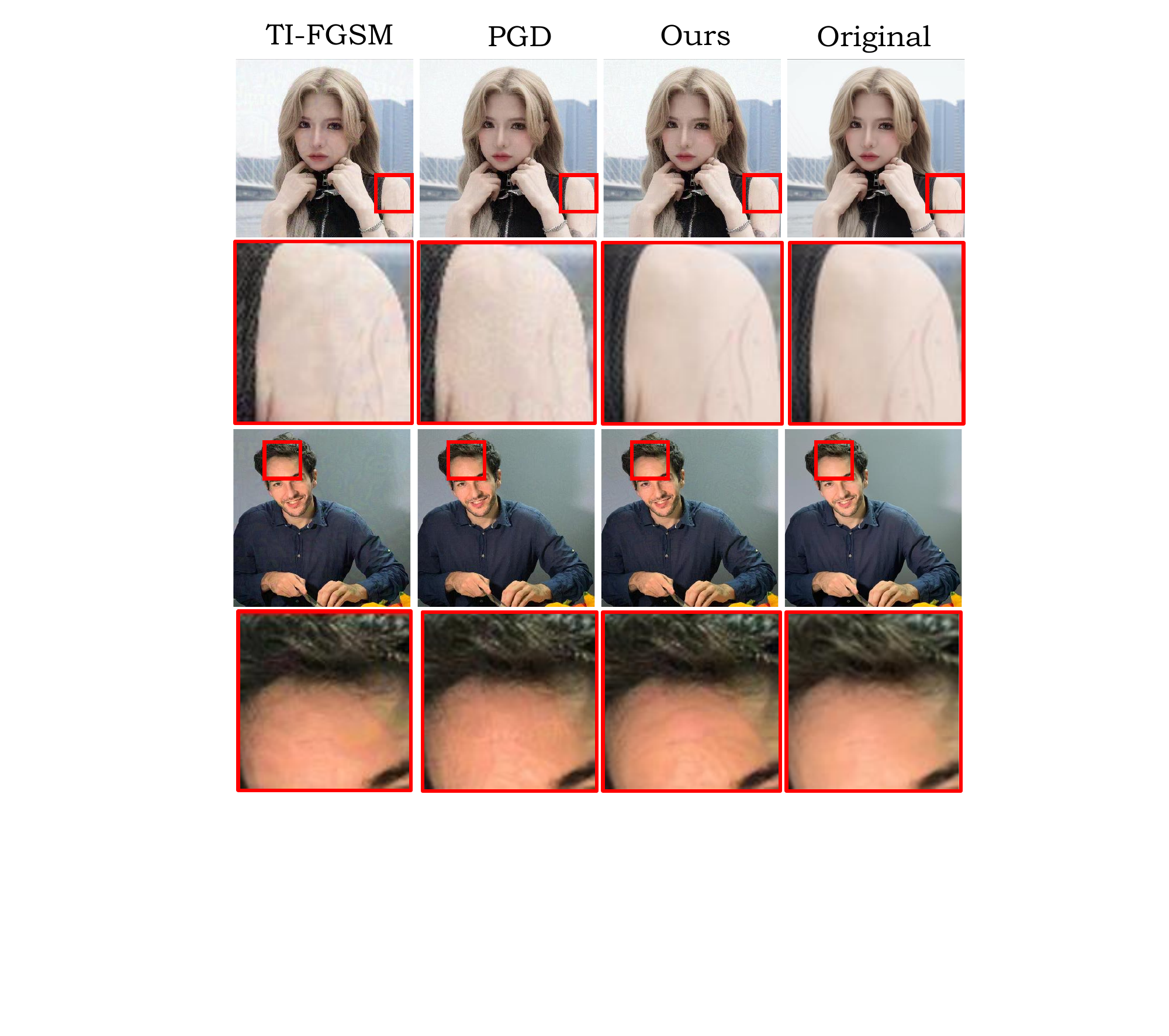}
    \caption{Qualitative comparison with representative equal-budget baselines. Insets show enlarged regions. Our protected images retain local detail with fewer visible artifacts.}
    \label{fig:qual}
\end{figure}

Figure~\ref{fig:qual} compares representative protected images and enlarged regions. Our results remain visually closer to the clean images than TI-FGSM and PGD, consistent with the higher PSNR and SSIM in Table~\ref{tab:image_protection}.

\subsubsection{Black-Box Transfer}

\begin{table}[t]
    \centering
    \caption{Strict black-box transfer to Yo'LLaVA under $\epsilon=4/255$. Each row excludes the listed surrogate from the six-encoder pool and optimizes over the remaining five; the target encoder is never used. Entries are protection rates (\%); higher is better.}
    \vspace{-.5em}
    \label{tab:blackbox_baselines_grid}
    \setlength{\tabcolsep}{3pt}
    \renewcommand{\arraystretch}{1}
    \begin{tabular}{@{}lccccc@{}}
        \toprule
        Excluded surrogate
        & I-FGSM
        & PGD
        & MI-FGSM
        & TI-FGSM
        & Ours \\
        \midrule
        CLIP-B/16
        & 24.4 & 28.9 & 28.9 & 6.7 & \textbf{32.2} \\
        CLIP-L/14-224
        & 10.0 & 12.2 & \textbf{18.9} & 5.6 & 8.9 \\
        CLIPA-L/14-336
        & 22.2 & 26.7 & 23.3 & 7.8 & \textbf{27.8} \\
        SigLIP
        & 30.0 & 31.1 & 30.0 & 6.7 & \textbf{35.6} \\
        ConvNeXtV2
        & 30.0 & 30.0 & \textbf{31.1} & 7.8 & \textbf{31.1} \\
        Supervised ViT-L
        & 40.0 & 42.2 & 42.2 & 18.9 & \textbf{48.9} \\
        \midrule
        \myrowcolor Average
        & 26.1 & 28.5 & 29.1 & 8.9 & \textbf{30.7} \\
        \bottomrule
    \end{tabular}
\end{table} 

\begin{table}[t]
    \centering
    \caption{\small Prompt robustness at $\epsilon=4/255$: protection rates (\%) across five equivalent identity-presence prompts.}
    \label{tab:prompt_invariance}
    \setlength{\tabcolsep}{6pt}
    \renewcommand{\arraystretch}{1.08}
    \begin{tabular}{@{}lccccc@{}}
        \toprule
        Method & P1 & P2 & P3 & P4 & P5 \\
        \midrule
        \myrowcolor Ours & 91.0 & 92.0 & 92.0 & 93.0 & 86.0 \\
        \bottomrule
    \end{tabular}
\end{table}

\noindent\textbf{Prompt variation.}
As shown in Table~\ref{tab:prompt_invariance}, across five semantically equivalent subject-presence prompts, protection ranges from $86.0\%$ to $93.0\%$, indicating limited sensitivity to prompt wording.

\begin{table}[t] 
\centering 
\caption{VQA-style identity-query protection on MyVLM under $\epsilon=4/255$. Entries are protection rates (\%), averaged over three prompts per image; higher is better.} 
\label{tab:vqa} 
\setlength{\tabcolsep}{3.2pt} \renewcommand{\arraystretch}{1.08} \begin{tabular}{@{}lccccc@{}} \toprule Method & Khanhvy & Viruss & Thao & Yuheng & Mean \\ \midrule FGSM & 6.7 & 36.7 & 43.3 & 33.3 & 30.0 \\ I-FGSM & 10.0 & 60.0 & 50.0 & 13.3 & 33.3 \\ MI-FGSM & \textbf{13.3} & 63.3 & 46.7 & 20.0 & 35.8 \\ TI-FGSM & \textbf{13.3} & 63.3 & 43.3 & 40.0 & 40.0 \\ PGD & \textbf{13.3} & 66.7 & 50.0 & 20.0 & 37.5 \\ \midrule 
\myrowcolor Ours & 10.0 & \textbf{73.3} & \textbf{53.3} & \textbf{46.7} & \textbf{45.8} \\ \bottomrule 
\end{tabular} 
\end{table}

\begin{figure*}[t]
    \centering

    \begin{minipage}[t][5.5cm][t]{0.58\textwidth}
        \vspace{0pt}
        \centering
        \captionsetup{position=top}
        \captionof{table}{
            JPEG compression robustness on Yo'LLaVA at $\epsilon=4/255$,
            averaged across identities. Protected images are saved with
            JPEG quality $Q$ using chroma subsampling setting \texttt{ss0}.
            Entries report protection rate (\%); higher is better.
        }
        \label{tab:jpeg_compression_robustness}

        \vfill

        {\small
        \setlength{\tabcolsep}{2.5pt}
        \renewcommand{\arraystretch}{1.12}
        \begin{tabular}{@{}lccccc@{}}
            \toprule
            \textbf{$Q$}
            & \textbf{FGSM}~\cite{goodfellow2015explaining}
            & \textbf{I-FGSM}~\cite{kurakin2017adversarial}
            & \textbf{MI-FGSM}~\cite{dong2018boosting}
            & \textbf{PGD}~\cite{madry2018towards}
            & \textbf{Ours} \\
            \midrule
            95 & 5.6 & 54.4 & 58.9 & 52.2 & \textbf{86.7} \\
            85 & 5.6 & 13.3 & 27.8 & 10.0 & \textbf{37.8} \\
            75 & 5.6 & 6.7  & 12.2 & 6.7  & \textbf{13.3} \\
            65 & 4.4 & 5.6  & \textbf{10.0} & 6.7 & 8.9 \\
            \bottomrule
        \end{tabular}\par}
        \vspace{0pt}
    \end{minipage}
    \hfill
    \begin{minipage}[t][6.0cm][t]{0.39\textwidth}
        \vspace{0pt}
        \centering
        \captionsetup{position=top}
        \captionof{figure}{
            Sensitivity to the number of reference images used to estimate
            the identity prototype. Protection improves as additional
            references are added.
        }
        \label{fig:num_refs}

        \vfill

        \includegraphics[width=\linewidth]{
            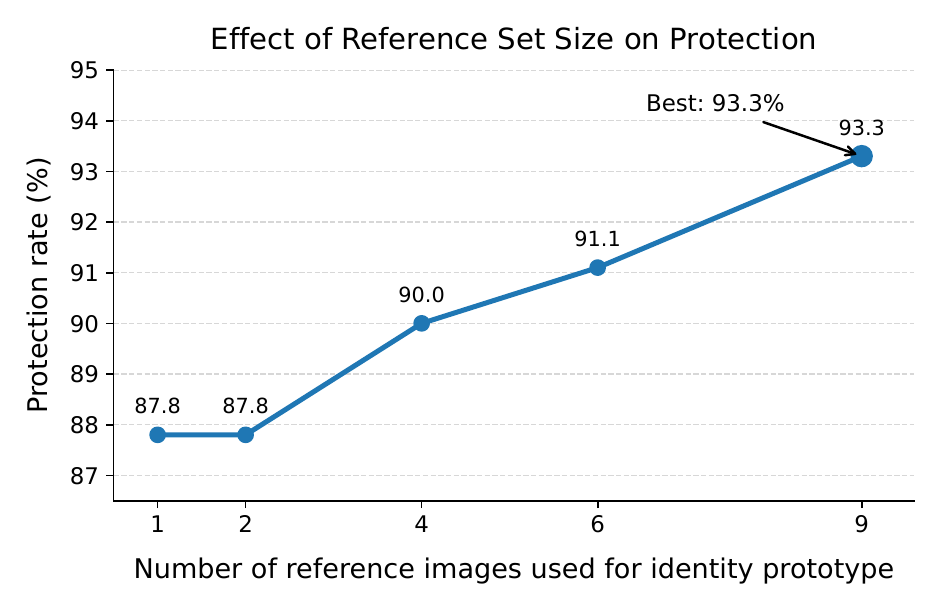
        }\par
        \vspace{0pt}
    \end{minipage}
\end{figure*}

Table~\ref{tab:image_protection} reports reactive protection. Our method achieves protection rates of $93.3\%$ on Yo'LLaVA and $71.1\%$ on MyVLM, exceeding the strongest equal-budget baselines by $5.5$ and $11.1$ percentage points, respectively. When the unrestricted references are included, it remains $5.5$ points above the strongest listed baseline on both models. The full method also provides the highest visual fidelity, with $39.30$\,dB PSNR and $0.958$ SSIM. 
The clean row is important when interpreting these rates: the unprotected failure rate is $3.3\%$ on Yo'LLaVA but $50.0\%$ on MyVLM. Relative to clean, our method increases recognition failure by $90.0$ and $21.1$ percentage points, respectively.

Table~\ref{tab:proactive_results} evaluates proactive protection. Our method achieves the highest average protection rate of $95.0\%$, exceeding the strongest reported baseline, TI-FGSM at $87.0\%$, by $8.0$ percentage points. It obtains at least $70.0\%$ protection for every identity and reaches $100\%$ for eight of the ten identities. These results show effective prevention of identity binding across the evaluated subjects.

We evaluate strict black-box transfer using the equal-budget feature-space baselines, for which encoder substitution is directly defined. The victim-aligned CLIP ViT-L/14@336 encoder is never used during protection generation. For each row of Table~\ref{tab:blackbox_baselines_grid}, one surrogate is excluded and the perturbation is optimized over the remaining five before evaluation on Yo'LLaVA.

Our method achieves an average protection rate of $30.7\%$, compared with $29.1\%$ for the strongest baseline, MI-FGSM. Although absolute black-box performance remains below the gray-box results, these findings show transfer without access to the target encoder.

\subsubsection{Robustness and Sensitivity Analyses}

We examine sensitivity to prompt wording, generalization to personalized VQA queries, and the number of images used to estimate the identity prototype.

\noindent\textbf{VQA-style identity queries.}
Table~\ref{tab:vqa} evaluates three yes/no personalized VQA prompts per image across four MyVLM identities. Our method achieves the highest mean protection rate of $45.8\%$, exceeding TI-FGSM at $40.0\%$ by $5.8$ percentage points and obtaining the best result for three of the four identities.

\noindent\textbf{JPEG compression robustness.}
Table~\ref{tab:jpeg_compression_robustness} evaluates protection after JPEG recompression across four quality factors at $\epsilon=4/255$. At $Q=95$, our method retains 86.7\% protection, outperforming the strongest baseline, MI-FGSM, by 27.8 percentage points. It remains strongest at $Q=85$ and $Q=75$, achieving 37.8\% and 13.3\%, respectively. Under severe compression at $Q=65$, all methods fall to 10.0\% or below, with our method remaining competitive at 8.9\%. Overall, our
method demonstrates greater JPEG robustness than the evaluated
baselines, consistent with its low-frequency design.

\noindent\textbf{Number of reference images.}
Figure~\ref{fig:num_refs} shows that protection increases from $87.8\%$ with one or two references to $90.0\%$, $91.1\%$, and $93.3\%$ with four, six, and nine references, respectively. This $5.5$-point improvement is consistent with the motivation for aggregating identity information across multiple images.

\section{Conclusion}

We introduced an image-level defense against unauthorized identity recognition in personalized LVLMs. Anti-Persona aggregates multiple reference embeddings into an identity prototype and optimizes protected images to disrupt identity-consistent encoder features. Spatial and frequency-domain refinements are designed to preserve visual quality and improve resilience to image compression. Experiments show effective reactive protection on Yo'LLaVA and MyVLM and proactive protection on Yo'LLaVA, while maintaining high visual fidelity. The method remains stable across prompt formulations, extends to VQA-style identity queries, and improves black-box transfer under encoder mismatch. These results motivate stronger privacy safeguards for personalized multimodal systems.

\bibliographystyle{assets/plainnat}
\bibliography{resources/main}

\clearpage
\beginappendix   

\setcounter{table}{0}
\setcounter{figure}{0}
\setcounter{equation}{0}
\renewcommand{\thetable}{A\arabic{table}}
\renewcommand{\thefigure}{A\arabic{figure}}
\renewcommand{\theequation}{A\arabic{equation}}

\section{Ablation Studies}
\label{app:ablations}

All ablations use reactive Yo'LLaVA in the victim-aligned gray-box setting with
CLIP ViT-L/14@336, $\epsilon=4/255$, and seed 42. Protection rate (PR) is
measured over 90 pooled queries across 10 identities, and PSNR is computed
between the clean and protected images.

\subsection{Protection Objective and Perturbation Refinements}
\label{app:abl_components}

We use the loss definitions and weights from \cref{sec:method}.
Table~\ref{tab:app_abl} varies the active loss terms and perturbation
refinements. The loss-only rows omit both refinements, while the refinement
rows use the complete objective. The full configuration attains the highest
observed protection rate, $93.3\%$, with a PSNR of $39.30$\,dB.

\begin{table}[H]
\centering
\caption{Component ablation on reactive Yo'LLaVA in the victim-aligned
gray-box setting (CLIP ViT-L/14@336, $\epsilon=4/255$). PSNR is in dB.
``Smoothing'' denotes masked Gaussian smoothing of the update, and ``DCT''
denotes low-pass projection of the perturbation. Loss rows omit both
refinements; refinement rows use the complete objective.}
\label{tab:app_abl}
\small
\begin{tabular}{@{}lcc@{}}
\toprule
Configuration & PR (\%) $\uparrow$ & PSNR $\uparrow$ \\
\midrule
$\mathcal{L}_{\mathrm{target}}$                         & 35.6 & 38.09 \\
$\mathcal{L}_{\mathrm{self}}$                           & 90.0 & 37.89 \\
$\mathcal{L}_{\mathrm{id}}$                             & 75.6 & 38.36 \\
$\mathcal{L}_{\mathrm{id}}+\mathcal{L}_{\mathrm{self}}$ & 70.0 & 38.16 \\
\addlinespace
$\mathcal{L}_{\mathrm{all}}$ (no refinement)            & 90.0 & 38.29 \\
$\mathcal{L}_{\mathrm{all}}$ + smoothing                & 86.7 & 38.31 \\
$\mathcal{L}_{\mathrm{all}}$ + DCT                      & 87.8 & 39.31 \\
\myrowcolor
$\mathcal{L}_{\mathrm{all}}$ + smoothing + DCT \textbf{(Ours)}
                                                        & \textbf{93.3} & 39.30 \\
\bottomrule
\end{tabular}
\end{table}

\subsection{Refinement Hyperparameters}
\label{app:abl_hparams}

Table~\ref{tab:app_hparam} reports sensitivity to the Gaussian smoothing width
$\sigma$ and the DCT keep parameter $b$. For the triangular DCT mask, $b$
retains coefficients satisfying $u+v\leq b$ within each $8\times8$ block.
One parameter is varied while the other remains fixed. The selected setting
$(\sigma,b)=(0.8,5)$ provides the highest observed protection rate in these
sweeps. The nearby setting $b=4$ gives slightly higher PSNR with a small
reduction in protection rate.

\begin{table}[H]
\centering
\caption{Sensitivity to the Gaussian smoothing width $\sigma$ and DCT keep
parameter $b$ on reactive Yo'LLaVA at $\epsilon=4/255$. One parameter is
varied while the other remains at its selected value $(\sigma,b)=(0.8,5)$.
Bold denotes the selected setting.}
\label{tab:app_hparam}
\small
\setlength{\tabcolsep}{5pt}
\begin{tabular}{@{}lrrrrrrr@{}}
\toprule
$\sigma$ ($b=5$)     & 0.2   & 0.4   & 0.6   & \textbf{0.8}   & 1.0   & 1.2   & 1.6   \\
\midrule
PR (\%) $\uparrow$   & 87.8  & 90.0  & 86.7  & \textbf{93.3}  & 87.8  & 88.9  & 88.9  \\
PSNR (dB) $\uparrow$ & 39.32 & 39.32 & 39.33 & \textbf{39.30} & 39.34 & 39.34 & 39.33 \\
\midrule
$b$ ($\sigma=0.8$)   & 2     & 3     & 4     & \textbf{5}     & 6     & 7     & 8     \\
\midrule
PR (\%) $\uparrow$   & 73.3  & 78.9  & 92.2  & \textbf{93.3}  & 91.1  & 88.9  & 86.7  \\
PSNR (dB) $\uparrow$ & 39.84 & 39.71 & 39.54 & \textbf{39.30} & 39.10 & 38.88 & 38.72 \\
\bottomrule
\end{tabular}
\end{table}

\section{Computational Cost}
\label{app:cost}

Table~\ref{tab:app_cost} reports computational cost for the default 500-step
configuration. PGD and Anti-Persona use the same vision encoder, so their
dominant encoder computation per step is the same. The reported wall-clock and
memory measurements include all method-specific loss and refinement
operations. Prototype features are computed once per identity and reused when
protecting its images.

\begin{table}[H]
\centering
\caption{Computational cost on one NVIDIA RTX A6000 using CLIP ViT-L/14@336
and $\epsilon=4/255$. Wall-clock time and peak memory are measured from a
single run; encoder FLOPs are analytically estimated.}
\label{tab:app_cost}
\small
\begin{tabular}{@{}lcc@{}}
\toprule
                           & PGD     & \textbf{Ours} \\
\midrule
Time / image (500 steps)   & 75\,s   & 82\,s ($+9.3\%$) \\
Peak GPU memory            & 3.2\,GB & 3.7\,GB \\
Encoder TFLOPs / step      & 0.57    & 0.57 \\
Prototype build / identity & --      & 0.3\,s \\
\bottomrule
\end{tabular}
\end{table}

\end{document}